\documentclass[11pt]{article}

\usepackage[preprint]{acl}

\usepackage{times}
\usepackage{latexsym}
\usepackage{amsmath}
\usepackage{amssymb}
\usepackage{multirow}
\usepackage[table]{xcolor}
\usepackage[most]{tcolorbox}
\usepackage{xspace}
\usepackage{enumitem}
\usepackage[most]{tcolorbox}

\newtcolorbox{promptbox}[1]{
  breakable,
  enhanced,
  colback=gray!3,
  colframe=black!50,
  boxrule=0.6pt,
  arc=2pt,
  left=6pt,
  right=6pt,
  top=6pt,
  bottom=6pt,
  title=\textbf{#1},
  colbacktitle=black!60,
  coltitle=white,
  fonttitle=\bfseries,
  toptitle=4pt,
  bottomtitle=4pt,
  before skip=0.6em,
  after skip=0.8em
}

\setlist[itemize]{
leftmargin=1.3em,
itemsep=0.25em,
topsep=0.25em,
parsep=0em
}

\setlist[enumerate]{
leftmargin=1.4em,
itemsep=0.2em,
topsep=0.2em,
parsep=0em
}

\usepackage[T1]{fontenc}
\usepackage{makecell}
\usepackage[utf8]{inputenc}

\usepackage{microtype}

\usepackage{inconsolata}

\usepackage{graphicx}
\usepackage{booktabs}

\title{ChatHealthAI: Aligning Electronic Health Record Representations with Large Language Models for Grounded Clinical Reasoning}

\author{
Bo-Hong Wang$^{1,2}$,
Baicheng Peng$^{1}$,
Ruilin Wang$^{1,2}$,
Jun Bai$^{1,2}$,
Ziyang Song$^{1,2}$,
Yue Li$^{1}$\thanks{Corresponding author: yueli@cs.mcgill.ca} \\
$^{1}$School of Computer Science, McGill University, Montreal, QC, Canada\\
$^{2}$Mila - Quebec AI Institute, Montreal, QC, Canada
}

\begin{document}
\maketitle
\begin{abstract}
Large language models (LLMs) exhibit strong natural-language reasoning abilities for clinical decision support, but struggle to effectively model structured longitudinal electronic health records (EHRs). In contrast, EHR foundation models can learn predictive patient representations, yet lack interpretable language-based reasoning. 
To bridge this gap, we propose ChatHealthAI, a multimodal reasoning framework that aligns structured EHR representations from a pretrained EHR foundation model with the semantic space of a frozen LLM through a task-aware resampler. 
By integrating longitudinal patient representations with refined clinical event descriptions, ChatHealthAI enables clinically grounded natural-language reasoning while maintaining accurate patient prediction.
We evaluated ChatHealthAI on three clinical predictive tasks from the EHRSHOT benchmark. 
Results show that ChatHealthAI improves reasoning quality and interpretability while preserving competitive predictive performance. 
These findings highlight the potential of integrating EHR foundation models with pretrained LLMs for interpretable clinical prediction.
\end{abstract}
\section{Introduction}
Large Language Models (LLMs) have shown strong capabilities in medical question answering, clinical reasoning and healthcare decision support, making them increasingly appealing for clinical applications that require interpretable natural-language explanations~\citep{singhal2023large,singhal2025medpalm2, saab2024gemini, thirunavukarasu2023trialling, liu2024medicalllmsurvey}. However, clinical predictions often require not only output labels but also interpretable explanations grounded in patient records~\citep{lauritsen2019xai}. This is challenging for structured longitudinal electronic health records (EHRs),  which consist of coded clinical events whose temporal order and clinical context are important for prediction~\citep{pang2021cehrbert, choi2016retain}. Simply serializing EHR events into an LLM prompt can exceed context limits and lose temporal and structural patterns~\citep{wu2024llemr, pang2021cehrbert, saab2024gemini}. This motivates integrating LLM-based reasoning with representations learned from structured EHR trajectories.

EHR foundation models address this complementary modeling problem by pretraining on large-scale structured clinical trajectories~\citep{wornow2023shaky}. Models such as CLMBR-T-Base~\citep{EHRSHOT}, Med-BERT~\citep{rasmy2021medbert}, and BEHRT~\citep{li2020behrt} learn patient representations from longitudinal EHR data and have shown strong performance on downstream clinical prediction tasks. These models capture predictive patterns from diagnosis, medication, laboratory, and other clinical event sequences, but their outputs are typically latent embeddings or risk scores rather than natural-language explanations
~\citep{li2020behrt,rasmy2021medbert,EHRSHOT}.
This creates a methodological gap between predictive EHR representation learning and natural-language clinical reasoning. EHR embeddings and LLM token embeddings are learned from heterogeneous input spaces~\citep{alayrac2022flamingovisuallanguagemodel,Richard2024.04.30.591835, saab2024gemini, moor2023foundation, li2026llm4ehr} and training objectives, and therefore are not naturally aligned in a shared representation space. Consequently, EHR embeddings cannot be directly treated as meaningful LLM inputs. Effective clinical reasoning therefore requires explicit alignment between structured EHR representations and the LLM representation space~\citep{shoham2024cpllm,liao2025ehrr1}.


To address this challenge, we propose ChatHealthAI, a multimodal clinical reasoning framework that aligns structured EHR representations with a frozen open-source LLM to enable grounded clinical prediction and reasoning generation. Specifically, we use CLMBR-T-Base~\citep{EHRSHOT} as the EHR foundation model and Deepseek-R1-Distill-Qwen-14B as LLM.
ChatHealthAI aligns patient representations learned from structured EHR trajectories with a frozen LLM through a task-aware resampler~\citep{Richard2024.04.30.591835}.
It further incorporates selected clinical events as textual evidence, enabling the model to generate grounded clinical reasoning and task-specific prediction conclusions.
We evaluate ChatHealthAI on the EHRSHOT benchmark~\citep{EHRSHOT}, including length-of-stay (LOS), ICU admission, and 30-day readmission prediction. Results show that ChatHealthAI maintains competitive predictive performance and improves reasoning quality, highlighting EHR-LLM alignment as a promising direction for interpretable clinical prediction.

In summary, our contributions are:

\begin{itemize}\setlength\itemsep{0pt}

\item We propose ChatHealthAI, a multimodal clinical reasoning framework that aligns structured longitudinal EHR representations with frozen LLM to support clinically grounded prediction and natural-language reasoning.

\item We integrate task-aware latent EHR representations with refined clinical event evidence to improve clinically grounded and interpretable reasoning over longitudinal patient trajectories.

\item We evaluate ChatHealthAI on multiple clinical prediction tasks from the EHRSHOT benchmark, including LOS, ICU admission, and 30-day readmission. Experimental results demonstrate improved reasoning quality while maintaining competitive predictive performance.

\end{itemize}

\section{Related Work}
\label{sec:related_work}

\paragraph{EHR representation learning.}
Prior EHR foundation models mainly focus on learning predictive patient embeddings from structured EHR data. Transformer-based EHR foundation models such as CEHR-BERT~\citep{pang2021cehrbert}, Med-BERT~\citep{rasmy2021medbert}, and CLMBR-T-Base~\citep{EHRSHOT} learn longitudinal patient representations through large-scale pretraining on structured EHR trajectories. EHRSHOT provided a standard few-shot evaluation benchmark used to evaluate CLMBR-T-Base~\citep{EHRSHOT}. However, these models mainly focus on predictive representation learning and do not explicitly support grounded natural-language clinical reasoning generation. 

\paragraph{Clinical Large Language Models.}

Recent clinical LLMs have demonstrated strong performance in medical question answering, clinical reasoning, and healthcare decision support~\citep{singhal2023large,singhal2025medpalm2, liu2024medicalllmsurvey, thirunavukarasu2023trialling, zhu2025askpatients, saab2024gemini}. Recent surveys further highlight the growing adoption of LLMs in diagnosis support, medical summarization, healthcare communication, and patient education~\citep{liu2024medicalllmsurvey, jang-etal-2025-chatbot, aydin2024llm_patient_education, zhu2025askpatients}. Despite these advances, most clinical LLMs primarily operate on textual inputs and do not explicitly leverage structured longitudinal EHR representations, limiting their ability to fully capture temporal clinical patterns and patient trajectories.~\citep{singhal2023large, singhal2025medpalm2, moor2023foundation, shoham2024cpllm, liao2025ehrr1}.

\paragraph{Representation Alignment between Foundation Models and LLMs.}
Recent studies have explored aligning pretrained foundation model representations with LLMs to improve reasoning over non-text data~\citep{xiao2024palm2vadapter,li2023blip2,li2026llm4ehr,zhu2023minigpt4, liu2023llava}. Multimodal frameworks such as Flamingo~\citep{alayrac2022flamingovisuallanguagemodel} and ChatNT~\citep{Richard2024.04.30.591835} suggest that lightweight alignment modules, such as perceiver resampler, can help bridge pretrained non-text representations with frozen LLMs.


\section{ChatHealthAI}

\begin{figure*}[t]
  \centering
  \includegraphics[width=\textwidth]{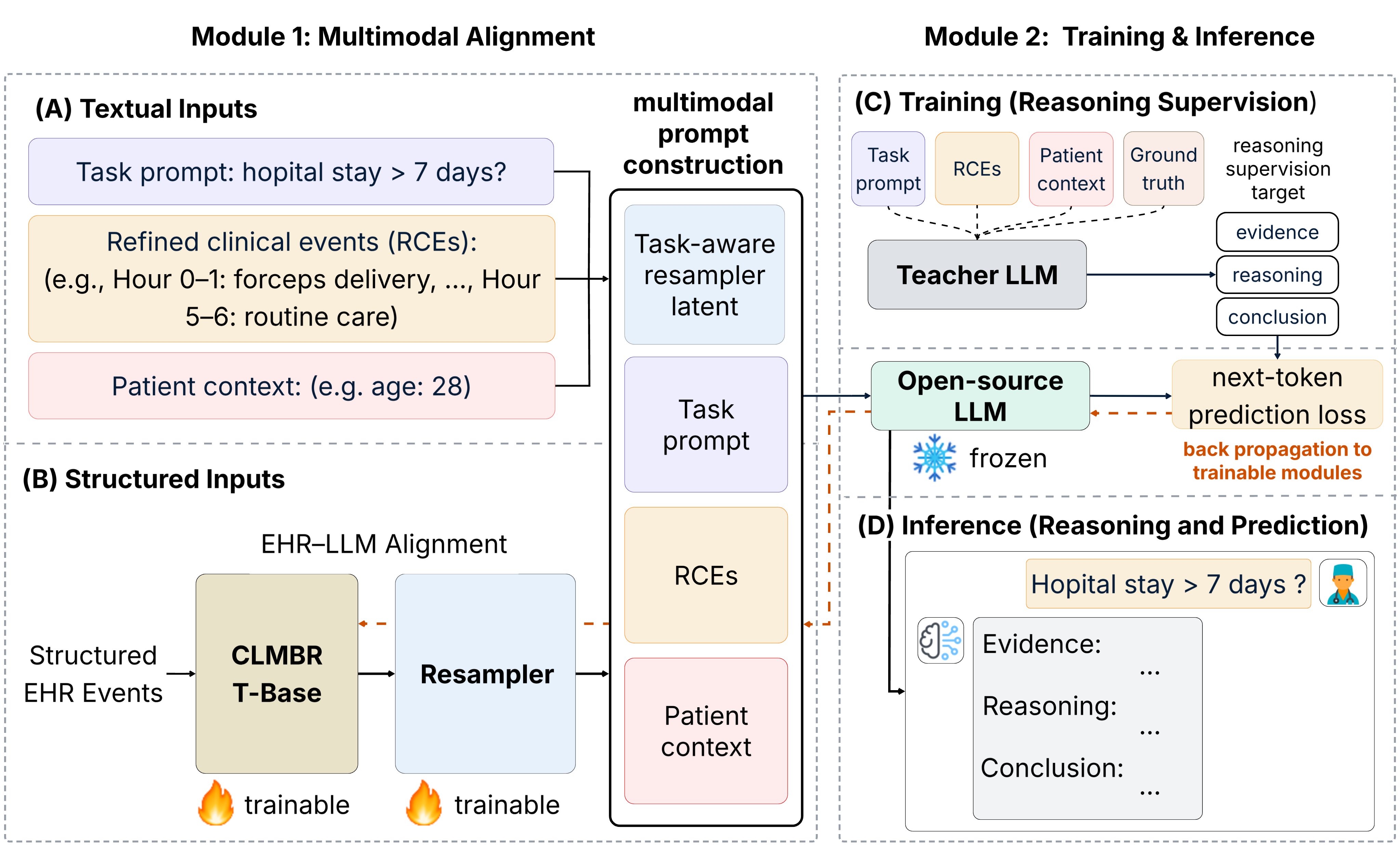}
\caption{
Overview of ChatHealthAI.
CLMBR-T-Base encodes structured EHR events into latent patient representations, which are aligned with a frozen open-source LLM through a task-aware resampler. The frozen open-source LLM additionally receives the task prompt, refined clinical events, and patient context as grounding information.
During training, teacher-generated reasoning targets supervise the EHR-side modules through next-token prediction loss.
During inference, the model generates evidence, reasoning, and a final prediction conclusion.
}
  \label{fig:overview}
\end{figure*}

We propose ChatHealthAI, a multimodal framework that aligns longitudinal EHR representations with a frozen open-source LLM to support clinical prediction with natural-language reasoning (Fig.~\ref{fig:overview}). ChatHealthAI encodes longitudinal patient trajectories with CLMBR-T-Base, maps the resulting representations to a frozen open-source LLM through a resampler, and uses refined clinical events as textual grounding for reasoning.

\subsection{Problem Formulation}
Let $\mathcal{E} = \{e_1, e_2, \ldots, e_T\}$ denote a patient's longitudinal EHR trajectory, where each event $e_t$ contains structured clinical information such as laboratory results, vital signs, medications, or diagnosis codes observed at timestamp $t$. Let $\mathcal{T}$ denote the textual inputs, including: task prompt, refined clinical events, patient context.

Given both the structured EHR trajectory $\mathcal{E}$ and the textual
inputs $\mathcal{T}$, ChatHealthAI aims to identify clinically relevant evidence from the longitudinal EHR trajectory, generate reasoning grounded in that evidence, and produce a prediction supported by both the structured EHR information and the reasoning process.

\subsection{Clinical Event Retrieval and Refinement}

Raw longitudinal EHR trajectories often contain lengthy, repetitive, and
low-information clinical events~\citep{rasmy2021medbert,EHRSHOT, myers2025rag}. Directly serializing the full EHR trajectory
into the LLM context is computationally inefficient and may dilute clinically
important signals within long sequences~\citep{Richard2024.04.30.591835, myers2025rag}. In addition, limited context windows
can lead to truncation of important events.

To address this problem, we select a compact subset of clinically
representative events from the longitudinal EHR trajectory for downstream
clinical reasoning. Specifically, we divide the EHR trajectory into temporal
chunks within a configurable lookback window (48 hours by default). We retrieve chunks using RAG ~\citep{rag-lewis,gao2024ragsurvey, zhao2025medrag, lu-etal-2024-clinicalrag, myers2025rag} with the retrieval query "Retrieve chunks that best summarize the patient's clinical course", and the retrieved chunks are
then refined by a LLM to select at most 30 representative clinical
events that best summarize the patient's trajectory~\citep{wright2025unstructured}. These \emph{refined clinical events} serve as textual grounding signals for downstream clinical reasoning~\citep{zhao2025medrag, lu-etal-2024-clinicalrag}.


\subsection{Task-Aware EHR-LLM Representation Alignment}

We encode the patient's event sequence $\mathcal{E}$ within the lookback
window using CLMBR-T-Base, obtaining contextualized EHR embeddings
$\mathbf{H}_{\text{CLMBR}} \in \mathbb{R}^{T \times d}$, where $T$ denotes
the number of clinical events and $d = 768$ is the EHR embedding dimension.
In our dataset, the number of clinical events varies substantially across
patients, with $T \in [100, 60000]$.

EHR embeddings and LLM token embeddings are learned from heterogeneous
input spaces and training objectives, and are therefore not naturally
aligned in a shared representation space. We initially explored using a
trainable linear projection layer to directly map CLMBR embeddings into the
LLM embedding dimension in order to bridge the representation gap between
EHR embeddings and LLM token embeddings. However, we found that simple linear projection was insufficient for the frozen LLM to effectively interpret longitudinal EHR representations,
leading to unstable reasoning generation and degraded predictive
performance. 

\begin{figure}
    \centering
    \includegraphics[width=1\linewidth]{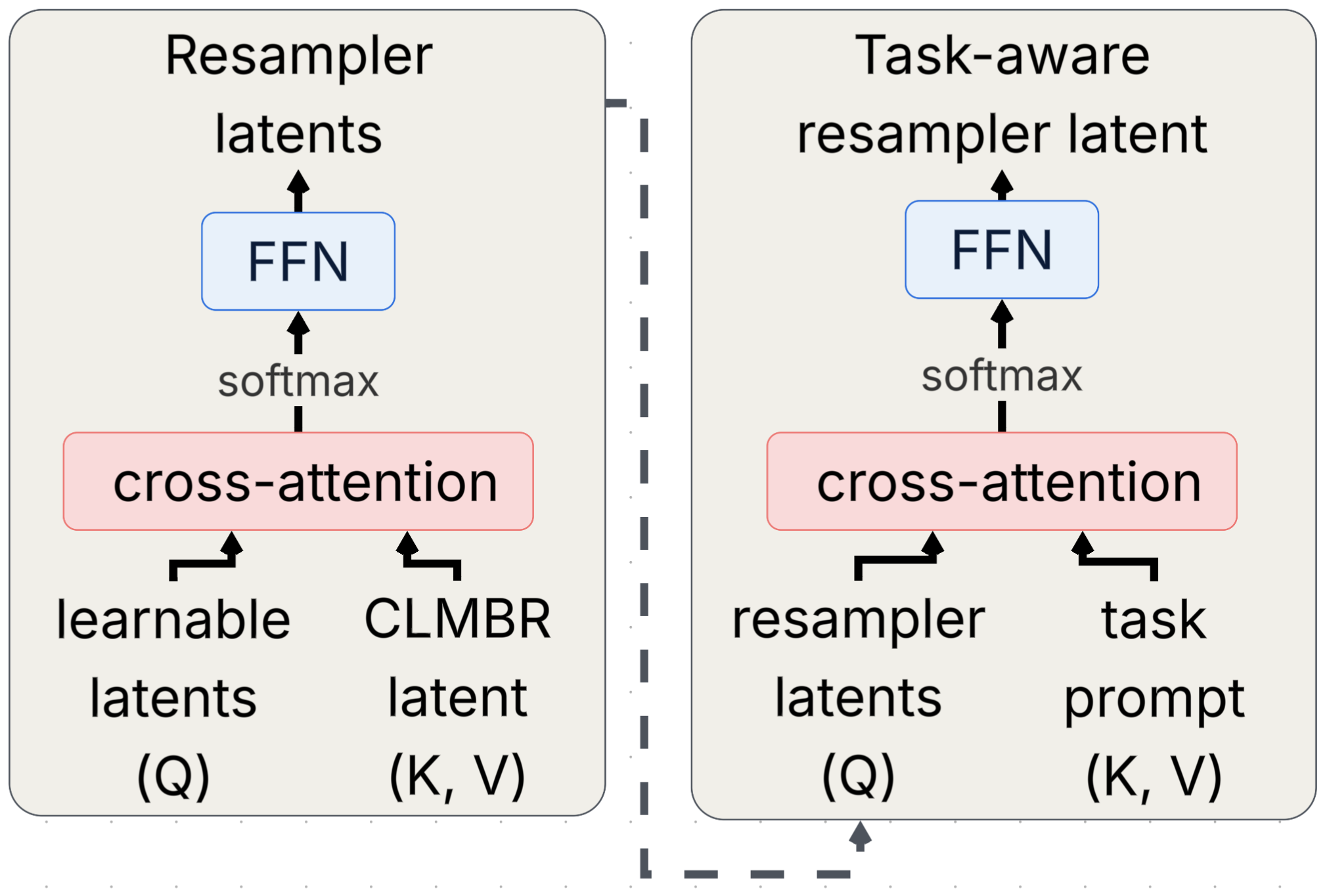}
    \caption{
Task-aware resampler.
Learnable latent queries first attend to CLMBR-T-Base embeddings to produce compact EHR latents, which then attend to the task prompt to generate task-aware representations.
}
    \label{fig:resampler}
\end{figure}

Therefore, we employ a perceiver resampler (Fig.~\ref{fig:resampler}), inspired by Flamingo~\citep{alayrac2022flamingovisuallanguagemodel} and ChatNT~\citep{Richard2024.04.30.591835}, to align structured EHR representations with the frozen LLM. The resampler contains $M{=}64$ learnable latent queries, $\mathbf{Q} \in \mathbb{R}^{M \times d}$, which attend to the CLMBR-T-Base embeddings through a cross-attention module followed by a feed-forward network (FFN)~\citep{vaswani2017attention, jaegle2021perceiver}: 
\begin{equation}
\mathbf{Z}
=
\mathrm{CrossAttn}(\mathbf{Q},\mathbf{K},\mathbf{V}),
\quad
\label{eq:resampler}
\end{equation}
where $\mathbf{K}$ and $\mathbf{V}$ are obtained from the CLMBR-T-Base embeddings,
$\mathbf{H}_{\text{CLMBR}} \in \mathbb{R}^{T \times d}$. The resampler compresses the EHR trajectory $\bf{H}$ from $T$ to a fixed set of M{=}64 latent tokens, thereby reducing computational cost and meeting the context length requirement. Then, we further condition the latent representation on a task-prompt embedding to support task-aware clinical reasoning. We apply another cross-attention layer followed by an FFN to produce a task-aware latent representation:
\begin{equation}
\mathbf{Z}'
=
\mathrm{CrossAttn}(\mathbf{Z},\mathbf{K},\mathbf{V}),
\label{eq:prompt_attn}
\end{equation}
where $\mathbf{K}$ and $\mathbf{V}$ are obtained from the task-specific
prompt embeddings, $\mathbf{P} \in \mathbb{R}^{L \times d}$, where $L$ is
the number of prompt tokens. This task-aware cross-attention enables the model to dynamically reweight latent resampler representations according to the target prediction task, thereby emphasizing different clinical information for different tasks.
\subsection{Clinical Reasoning Instruction Tuning}

The input to the frozen LLM consists of four components:
(1) the task prompt, (2) the task-aware latent representations, (3) the patient-level context such as  patient age at prediction time, and (4) serialized refined clinical events with relative timestamps measured from the first observed event. We use DeepSeek-R1-Distill-Qwen-14B as the frozen LLM backbone.


To supervise the generation of clinically grounded reasoning, we use GPT-oss-120B as a teacher model to generate structured reasoning targets.~\citep{wang2023selfinstruct,hsieh2023distilling, mukherjee2023orca} Given refined clinical events and the ground-truth, the teacher model produces structured answers consisting of: (1) clinical evidence, (2) step-by-step reasoning, and (3) a label-consistent conclusion.~\citep{hsieh2023distilling,mukherjee2023orca,wei2022chain}
These teacher-generated answers serve as supervision targets during training. Given a tokenized teacher-generated target sequence $y^* = (y^*_1, \ldots, y^*_N)$, we optimize the trainable ChatHealthAI modules based on the next-token prediction loss: 
\begin{equation} 
\mathcal{L}_{\mathrm{NTP}} = -\sum_{j=1}^{N} \log p_{\theta} \left( y^*_j \mid \mathbf{x}_{<j} \right), 
\label{eq:loss}
\end{equation}
where $\theta$ denotes the trainable parameters of CLMBR-T-base encoder and task-aware resampler, while the LLM backbone remains frozen. At inference time, ChatHealthAI autoregressively generates clinical evidence, reasoning, and a final prediction conclusion.

\section{Experimental Setting}
\subsection{Dataset}
We evaluated ChatHealthAI on \textbf{EHRSHOT} \citep{EHRSHOT}, which contains longitudinal structured EHR trajectories of diagnoses, medications, laboratory tests, procedures, etc. EHRSHOT provides  multiple standardized clinical prediction tasks, along with patient-level CLMBR-T-Base embeddings pre-trained on 2.57 million deidentified EHRs. We focused on three different clinical prediction tasks: (1) length-of-stay (LOS) prediction, which predicts whether the patient's hospitalization lasts at least 7 days, (2) ICU admission prediction, which predicts whether the patient will require intensive care unit admission, and (3) 30-day readmission prediction, which predicts whether the patient will be readmitted within 30 days after discharge. We also splited the train and test sets following the established EHRSHOT procedures. Table~\ref{tab:data} summarizes dataset statistics.

\begin{table}[h]
\centering\small
\caption{Dataset statistics for the EHRSHOT benchmark with three clinical prediction tasks. \textbf{Pos.} denotes the positive ratios (\%).}
\label{tab:data}
\begin{tabular}{lcccc}
\toprule
\textbf{Task} & \textbf{Pos.} & \textbf{Train} & \textbf{Val} & \textbf{Test} \\
\midrule
Length of Stay $\geq$ 7d & 25.3\% & 2569 & 2231 & 2195\\ 
ICU admission        & 4.5\% & 2402 & 2052 & 2037\\ 
30-day readmission   & 13.0\% & 2608 & 2206 & 2189\\ 
\bottomrule
\end{tabular}
\end{table}

\begin{table*}[t]
\centering
\small

\caption{
Main results on EHRSHOT benchmark test sets reported in
Precision / Recall / F1.
LLM-based methods output discrete labels;
CLMBR+Linear uses threshold~0.5.
Best results are highlighted in \textbf{bold}.
}

\label{tab:main}

\begin{tabular}{llccccccccc}
\toprule

\multirow{2}{*}{\textbf{Method}} &
\multirow{2}{*}{\textbf{Setting}} &
\multicolumn{3}{c}{\textbf{LOS}} &
\multicolumn{3}{c}{\textbf{ICU Admission}} &
\multicolumn{3}{c}{\textbf{Readmission}} \\

\cmidrule(lr){3-5}
\cmidrule(lr){6-8}
\cmidrule(lr){9-11}

&
&
P & R & F1 &
P & R & F1 &
P & R & F1 \\

\midrule

CLMBR + Linear
& --
& 0.667 & 0.286 & 0.401
& 1.0 & 0.01 & 0.02
& \textbf{0.670} & 0.110 & 0.190 \\

\multirow{2}{*}{Llama-3.1-8B-Instruct}
& zero-shot
& 0.253 & \textbf{0.996} & 0.403
& 0.05 & 1.0 & 0.09
& 0.119 & \textbf{0.987} & 0.213 \\

& finetuned
& 0.269 & 0.794 & 0.402
& 0.05 & 0.563 & 0.083
& 0.132 & 0.507 & 0.210 \\

\multirow{2}{*}{BioMistral-7B}
& zero-shot
& 0.252 & 0.976 & 0.398
& 0.05 & 1.0 & 0.09
& 0.120 & 0.981 & 0.214 \\

& finetuned
& 0.269 & 0.795 & 0.401
& 0.082 & 0.047 & 0.060
& 0.231 & 0.142 & 0.175 \\

\multirow{2}{*}{DeepSeek-R1-14B}
& zero-shot
& 0.258 & 0.993 & 0.409
& 0.046 & \textbf{0.835} & 0.087
& 0.120 & 0.969 & 0.214 \\

& finetuned
& 0.237 & 0.925 & 0.377
& 0.066 & 0.762 & 0.121
& 0.119 & 0.783 & 0.206 \\

\midrule

\textbf{ChatHealthAI}
& finetuned
& \textbf{0.442} & 0.551 & \textbf{0.491}
& \textbf{0.385} & 0.136 & \textbf{0.196}
& 0.208 & 0.242 & \textbf{0.224} \\

\bottomrule
\end{tabular}
\end{table*}

\subsection{Baselines}
We compared with CLMBR+Linear using frozen CLMBR-T-Base with a linear classifier~\citep{EHRSHOT} given a prediction threshold of 0.5, serving as a structured EHR prediction baseline without natural-language reasoning. We also included a prompt-based LLM baseline that directly uses serialized refined clinical events as input without using a resampler and an EHR representation. We evaluated four open-sourced models under zero-shot and finetuned settings: 3Llama-3.1-8B-Instruct~\citep{grattafiori2024llama3}, selected as an open-source instruction-tuned LLM; DeepSeek-R1-Distill-Qwen-14B~\citep{guo2025deepseekr1}, selected as a frozen LLM backbone of ChatHealthAI with  strong reasoning capabilities; BioMistral-7B~\citep{labrak2024biomistral}, selected as a pretrained backbone on large-scale biomedical textual data.

\subsection{Metrics}
In this work, our ChatHealthAI and baselines directly output structured natural-language outputs that include evidence, step-by-step reasoning, and conclusions, rather than providing prediction probabilities. We therefore used Precision, Recall, and F1 as the predictive metrics for evaluation. The prediction is obtained from the generated “Yes” or “No” token following ``Conclusion:'' in the output template. For CLMBR+Linear, which produces continuous prediction scores, we apply a fixed threshold of 0.5 to obtain prediction labels.
\subsection{Reasoning Evaluation}

We quantitatively evaluated the reasoning quality of each model using GPT-5.4, Claude Sonnet 4.5, 
and Gemini 2.5 Pro as independent judges under a structured rubric-based protocol \citep{liu-etal-2023-g, llm-judge, croxford2025automating}. Judges are instructed to use the task prompt, refined clinical events, and ground-truth, along with the model's generated explanation and predicted result for assessment. The explanation contains clinically relevant 
evidence, step-by-step reasoning, and a final conclusion, scored on eight criteria 
using a 1--5 rating scale.

\paragraph{Reasoning Quality.} We assessed the quality of each generated rationale across six dimensions: \textbf{(1) Evidence grounding} assesses whether claims are supported 
by the provided refined clinical events. \textbf{(2) Clinical relevance} measures whether the selected 
evidence is pertinent to the prediction task. \textbf{(3) Temporal reasoning} 
captures whether the explanation reflects progression, stability, escalation, or 
de-escalation over time. \textbf{(4) Clinical Coherence} evaluates whether the explanation follows clinically plausible reasoning. \textbf{(5) Completeness} evaluates whether sufficient evidence 
and reasoning support the conclusion. \textbf{(6) Safety of overclaiming} penalizes 
unsupported claims or clinically misleading statements~\citep{asgari2025clinical}.

\paragraph{Reasoning Utility.} We also assessed whether the explanation supports the final prediction in a clinically meaningful way via two criteria: \textbf{(1) Outcome alignment} measures whether the conclusion matches the 
ground-truth label, while \textbf{(2) Clinical usefulness} reflects whether the 
explanation aids a clinician in interpreting the model prediction.

\begin{table*}[t]
\centering
\footnotesize
\setlength{\tabcolsep}{4pt}

\caption{
Per-dimension LLM-judge reasoning quality scores on the LoS task
(1--5 scale, averaged over GPT-5.4, Claude Sonnet 4.5, and Gemini 2.5 Pro judges).
Column abbreviations --- Evid.\ Ground.: Evidence Grounding;
Clin.\ Relev.: Clinical Relevance;
Temp.\ Reason.: Temporal Reasoning;
Clin.\ Coher.: Clinical Coherence;
Compl.: Completeness;
Outcome Align.: Outcome Alignment;
Clin.\ Useful.: Clinical Usefulness.
}

\label{tab:reasoning}

\begin{tabular}{llcccccc|cc}
\toprule

\multirow{2}{*}{\textbf{Method}} &
\multirow{2}{*}{\textbf{Setting}} &
\multicolumn{6}{c|}{\textbf{Reasoning Quality}} &
\multicolumn{2}{c}{\textbf{Reasoning Utility}} \\

\cmidrule(lr){3-8}
\cmidrule(lr){9-10}

&
&
\thead{Evid.\\Ground.} &
\thead{Clin.\\Relev.} &
\thead{Temp.\\Reason.} &
\thead{Clin.\\Coher.} &
Compl. &
Safety &
\thead{Outcome\\Align.} &
\thead{Clin.\\Useful.} \\

\midrule

\multirow{2}{*}{Llama-3.1-8B-Instruct}
& zero-shot
& 3.655 & 3.823 & 1.766 & 3.088 & 2.540 & \textbf{3.148} & 2.153 & 2.268 \\

& finetuned
& 3.383 & 3.811 & 2.885 & 2.685 & 2.764 & 2.891 & 2.303 & 2.132 \\

\multirow{2}{*}{BioMistral-7B}
& zero-shot
& 2.710 & 3.055 & 1.307 & 2.043 & 1.478 & 2.443 & 1.862 & 1.474 \\

& finetuned
& 2.132 & 3.321 & 2.640 & 2.347 & 2.363 & 2.097 & 3.038 & 1.925 \\

\multirow{2}{*}{DeepSeek-R1-14B}
& zero-shot
& \textbf{3.638} & 4.057 & 3.160 & \textbf{3.233} & 3.263 & 2.826 & 2.170 & 2.318 \\

& finetuned
& 3.281 & 3.740 & 2.797 & 2.730 & 2.646 & 2.933 & 1.942 & 2.060 \\


\midrule

\textbf{ChatHealthAI}
& finetuned
& 3.463
& \textbf{4.165}
& \textbf{3.577}
& 3.195
& \textbf{3.379}
& 2.778
& \textbf{3.566}
& \textbf{2.817} \\

\bottomrule
\end{tabular}
\end{table*}

\section{Results}

\subsection{Main Prediction Results}

Prompting-only LLM baselines, especially in the zero-shot setting, generally achieve high recall but low precision (Table~\ref{tab:main}). This result indicates that prompting-only LLMs tend to over-predict positive outcomes when exposed to high-risk clinical events such as severe diagnoses, medications, or procedures. In some cases, zero-shot models predict nearly all samples as positive, resulting in an poor precision--recall trade-off. Instruction fine-tuning mitigates this issue but remains inconsistent across tasks. For example, the finetuned BioMistral-7B improves precision on readmission prediction while substantially reducing recall. In contrast, ChatHealthAI achieves a more balanced precision-recall trade-off and obtains the best F1 scores on all evaluated tasks among the compared methods. These results suggest that incorporating longitudinal EHR representations helps capture latent temporal and clinical patterns that cannot be reliably inferred from refined clinical events alone, thereby improving clinical risk discrimination beyond prompting-only LLMs.

\subsection{Reasoning Ability}













\begin{figure*}[t]
    \centering
    \includegraphics[width=1\linewidth]{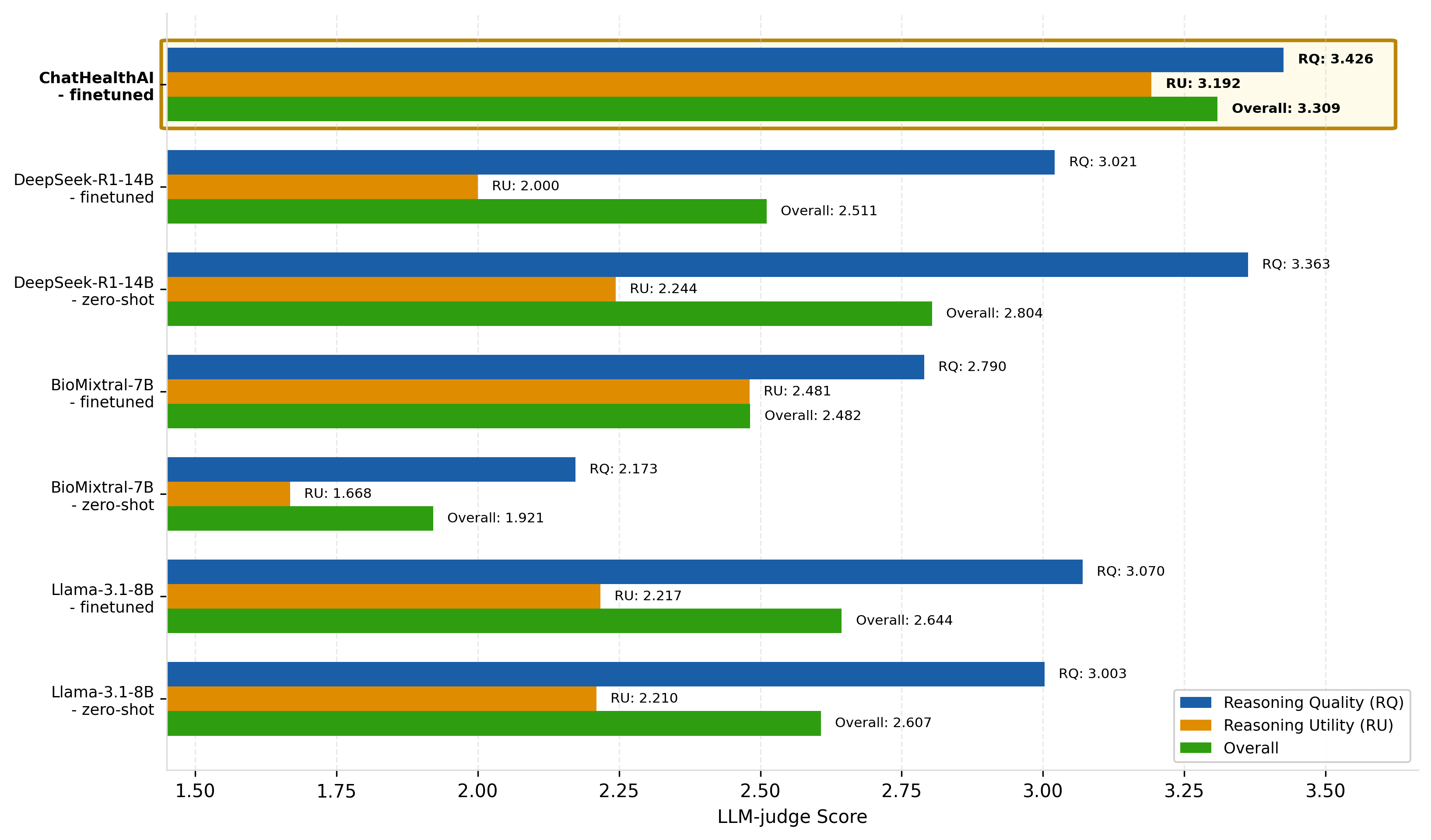}
    \caption{Average LLM-judges evaluation results on the length-of-stay prediction task. ChatHealthAI achieves the highest reasoning quality (RQ), reasoning utility (RU), and overall score among all compared baselines.}
    \label{fig:overall reasoning score}
\end{figure*}

ChatHealthAI shows improved reasoning performance across multiple criteria, particularly in clinical relevance, temporal reasoning, completeness, outcome alignment, and clinical usefulness (Table~\ref{tab:reasoning}). These improvements also lead to the highest overall reasoning performance, including average Reasoning Quality (RQ) score across six reasoning dimensions and average Reasoning Utility (RU) score across two reasoning utility criteria (Fig.~\ref{fig:overall reasoning score}).
Directly finetuning DeepSeek-R1-14B generally leads to lower reasoning quality and reasoning utility than the zero-shot baseline. This suggests that instruction finetuning alone may encourage the model to rely on isolated clinical events or salient medical terms, while failing to consistently capture the patient’s longitudinal EHR trajectory.

In contrast, ChatHealthAI shows consistent improvement in both Reasoning Quality and Reasoning Utility.  Its gains in temporal reasoning suggest that aligned longitudinal EHR representations help the model reason more effectively about clinical events over time. This improvement is particularly important for longitudinal clinical prediction tasks, where the significance of clinical events often depends on their temporal progression and relationships with other events. ChatHealthAI also improves completeness and clinical relevance, indicating that longitudinal EHR priors help generate explanations that are more grounded in the patient-specific clinical trajectory. Overall, these results suggest that longitudinal EHR representations provide useful temporal signals for improving clinical reasoning about quality utility beyond prompting-only LLMs.

\subsection{Ablation Study}
\begin{table}[t]
\centering
\footnotesize
\setlength{\tabcolsep}{5pt}
\caption{Ablation results on the LoS task.}
\label{tab:ablation}
\begin{tabular}{lccccc}
\toprule
\textbf{Variant} & \textbf{P} & \textbf{R} & \textbf{F1} & \textbf{RQ} & \textbf{RU} \\
\midrule
Full ChatHealthAI          & 0.442 & 0.551 & 0.491 & 3.665 & 3.425 \\
Random CLMBR            & 0.361 & 0.350 & 0.355 & 3.426 & 3.192 \\
w/ linear proj.    & N/A & N/A & N/A & 1.733 & 1.726 \\
\bottomrule
\end{tabular}
\end{table}

 Using the LOS task as an example, replacing CLMBR-T-Base embeddings with random embeddings reduces F1-score from 0.491 to 0.355 (Table~\ref{tab:ablation}), indicating that the clinical priors learned by CLMBR-T-base provide essential predictive information that cannot be recovered from refined clinical events alone. As expected, when CLMBR-T-Base embeddings are replaced with random vectors, the reasoning quality and clinical utility scores remain relatively high. This is because the reasoning quality is mainly supported by frozen LLMs and refine clinical events, whereas latent EHR representations are more important for prediction discrimination. However, compared with the zero-shot and finetuned DeepSeek-R1-14B backbone used in ChatHealthAI, the full ChatHealthAI framework has a more balanced recall-precision trade-off, and improved reasoning quality (RQ) and reasoning utility (RU). This observation suggests that, even without meaningful longitudinal EHR embeddings, the resampler can still learn to calibrate reasoning behavior and improve prediction.
In contrast, replacing the resampler with a simple linear projection layer causes the model to score very low on both RQ and RU, and results in invalid structured outputs that cannot be used for classification metric calculation. This indicates that linear layer does not compress and align CLMBR-T-Base’s embeddings into semantically meaningful latent tokens  for the frozen LLM. These results show that the resampler is not merely a dimensional projection module, but a critical alignment module for connecting structured EHR representations with LLM-based reasoning.

\begin{figure*}[t!]
    \centering
    \includegraphics[width=0.97\textwidth]{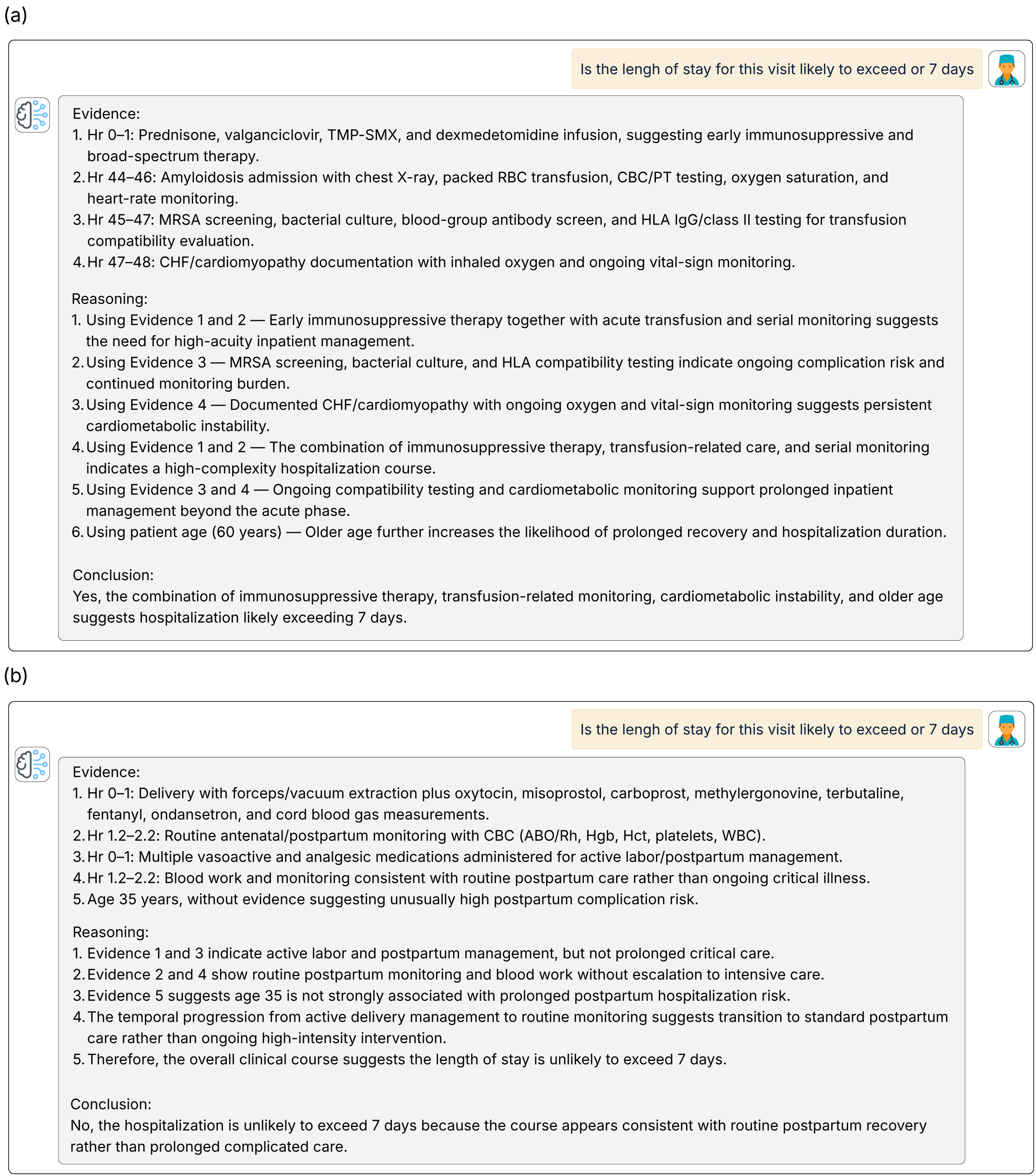}
    \caption{Case studies of ChatHealthAI-generated clinical reasoning. (a) A positive case involving immunosuppressive therapy and cardiometabolic instability. (b) A negative case involving routine postpartum recovery.}
    \label{fig:case_study_positive}
\end{figure*}

\section{Case Studies}
A representative positive LOS case involves immunosuppressive therapy and broad-spectrum therapy, followed by transfusion-related care, infection-related testing, and continued cardiometabolic monitoring across different stages of the 48-hour EHR timeline (Figure~\ref{fig:case_study_positive}a). ChatHealthAI organizes these events into a structured explanation. It first identifies early immunosuppressive therapy and transfusion as evidence of high-acuity inpatient management. It then connects these findings with later MRSA screening, bacterial culture, HLA compatibility testing, and ongoing vital-sign monitoring, suggesting continued complication risk and persistent clinical instability. Together, these temporally distributed findings suggest sustained treatment burden and failure to achieve early clinical stabilization. Finally, the model combines these clinical signals with the patient's age of 60 years to support the conclusion that the hospitalization is likely to exceed 7 days. The case illustrates the intended behavior of ChatHealthAI by showing how it links multiple clinical events across the longitudinal timeline and produces a reasoning chain consistent with the final prediction. 

A representative negative LOS case reflects routine postpartum recovery without escalation to intensive treatment (Figure~\ref{fig:case_study_positive}b). Although the patient received active delivery-related management during the early stage of hospitalization, the later clinical events consisted mainly of routine postpartum monitoring, blood work, and standard supportive care. ChatHealthAI correctly identifies the patient's transition from active
delivery management to stable postpartum recovery and concludes that this hospitalization is unlikely to exceed 7 days. This example further demonstrates that ChatHealthAI can distinguish between temporary intensive treatment and persistent clinical instability.

\section{Conclusion}
We developed ChatHealthAI, a multimodal clinical reasoning framework that aligns a structured longitudinal EHR representation with a frozen LLM to generate grounded clinical predictions and reasoning. Across three EHRSHOT prediction tasks, ChatHealthAI achieves a more balanced precision—recall trade-off and obtains the best F1 score on LOS and ICU admission prediction.  Reasoning evaluation demonstrates that ChatHealthAI achieves the best overall Reasoning Quality and Reasoning Utility scores. Our ablation study shows that the resampler is an essential alignment module, enabling the frozenLLM to effectively leverage structured EHR representations for reasoning and prediction. Overall, ChatHealthAI demonstrates the potential of integrating EHR foundation models with LLMs to build clinical prediction systems that are both predictive and interpretable.

\section*{Limitations}

Our proposed ChatHealthAI has several limitations. First, ChatHealthAI is evaluated only on the EHRSHOT benchmark, and its generalizability to other EHR datasets and healthcare systems remains uncertain. Second, reasoning quality is evaluated primarily using LLM-based judges rather than expert human clinicians. Although recent studies suggest that LLM judges can provide relatively consistent evaluation, they may not fully reflect real-world clinical preferences or expert-level reasoning assessment. Third, the reasoning supervision used for instruction finetuning is generated by a teacher LLM rather than expert clinicians. The generated explanations may still contain factual inaccuracies, incomplete reasoning, or clinically imperfect interpretations. Finally, limitations in framework compatibility between CLMBR-T-Base and newer open-source LLM architectures currently restrict evaluation on more recent frozen LLM backbones.


\bibliography{custom}

\clearpage
\appendix
\footnotesize

\section{Appendix}

\subsection{Prompt Templates}

\subsubsection{Clinical Event Refinement Prompt}

\begin{promptbox}{Clinical Event Refinement Prompt}

\textbf{You are selecting clinically important events for downstream reasoning.}

\textbf{Input:} Candidate temporal EHR chunks retrieved by retrieval-augmented retrieval.

\textbf{Task:} From the input chunks, select a compact but balanced set of clinically important events that best represents the patient's clinical course around prediction time.

\textbf{Strict constraints}
\begin{itemize}
\item Output at most 30 events total.
\item Preserve a balanced view of the case.
\item Include events suggesting:
\begin{itemize}
\item high complexity, severe illness, invasive care, or major intervention
\item routine care, stable course, uncomplicated recovery, or lower clinical intensity
\end{itemize}
\end{itemize}

\textbf{Selection rules}
\begin{itemize}
\item Prefer events that characterize the overall clinical course.
\item Prefer diversity over redundancy.
\item Preserve diagnoses, procedures, medications, care setting, monitoring, imaging, and clinically meaningful notes.
\item Keep routine or low-intensity events if they help distinguish routine versus complex clinical courses.
\item Remove only exact duplicates or clearly redundant repetitions.
\item Do not bias toward severe or critical events only.
\end{itemize}

\textbf{Event rules}
\begin{itemize}
\item Only use event names already appearing in the input.
\item Do not paraphrase event names.
\item Do not merge events.
\item Preserve original timestamps.
\item Group events under their original timestamps.
\end{itemize}

\end{promptbox}

\subsubsection{Reasoning Supervision Generation Prompt}

\begin{promptbox}{Reasoning Supervision Generation Prompt}

\textbf{You are generating reasoning supervision for a hospital length-of-stay task.}

\textbf{Task:}
\begin{itemize}
\item The question is whether this hospital visit will have a total length of stay of at least 7 days.
\item The label is already known and must be followed.
\item The goal is to generate structured reasoning supervision consistent with the known label.
\end{itemize}

\textbf{Output format}

The generated answer must follow the structure:

\begin{quote}
Evidence: 1) ... 2) ...

Reasoning:

Step 1: Using Evidence ...

Step 2: Using Evidence ...

Conclusion: Yes/No, ...
\end{quote}

\textbf{Generation rules}
\begin{itemize}
\item Use only information supported by the provided clinical events.
\item Do not hallucinate diagnoses, procedures, medications, or complications.
\item Each reasoning step must explicitly reference Evidence items.
\item Temporal progression, persistence, escalation, or lack of improvement must be incorporated into the reasoning process.
\item Avoid unsupported severity claims.
\item Use cautious clinical language such as ``suggests'', ``consistent with'', or ``likely'' when appropriate.
\item Conclusions must be based on treatment intensity, monitoring burden, procedures, and overall clinical complexity rather than explicit discharge timing.
\end{itemize}

\end{promptbox}

\subsubsection{LLM Judge Prompt}

\begin{promptbox}{LLM Judge Prompt}
\raggedright

\textbf{You are evaluating a structured clinical reasoning explanation generated by an AI model.}

\vspace{0.5em}

\textbf{You will be given:}
\begin{enumerate}
\item Task instruction.
\item Patient clinical events.
\item Ground-truth label.
\item Model prediction.
\item Model-generated explanation.
\end{enumerate}

\textbf{Expected structure:}
\begin{enumerate}
\item Evidence: a numbered list of clinical evidence items.
\item Reasoning: step-by-step reasoning that refers back to the evidence.
\item Conclusion: a final Yes/No prediction with a short explanation.
\end{enumerate}

\textbf{Evaluate all dimensions on a 1--5 Likert scale.}

\vspace{0.5em}

\textbf{General scoring guide}
\begin{itemize}
\item \textbf{1 = Poor:} mostly incorrect, unsupported, incoherent, or misleading.
\item \textbf{2 = Weak:} contains major problems, but a small portion is acceptable.
\item \textbf{3 = Fair:} partially correct and usable, but with clear limitations.
\item \textbf{4 = Good:} mostly correct, grounded, and useful, with minor issues.
\item \textbf{5 = Excellent:} fully grounded, clinically coherent, complete, and useful.
\end{itemize}

\vspace{0.5em}

\textbf{Dimensions}

\vspace{0.3em}

\textbf{1. evidence\_grounding}

Does each Evidence item and major reasoning claim come from the provided clinical events or patient context?

\begin{itemize}
\item 1 = mostly hallucinated or unsupported.
\item 2 = several important unsupported claims.
\item 3 = partially grounded, but some claims are vague or weakly supported.
\item 4 = mostly grounded, with only minor unsupported or vague claims.
\item 5 = all major claims are clearly grounded in the provided events.
\end{itemize}

\textbf{2. clinical\_relevance}

Is the selected evidence relevant to the task instruction?

\begin{itemize}
\item 1 = mostly irrelevant evidence.
\item 2 = limited relevance; many selected events do not help answer the task.
\item 3 = mixed relevant and irrelevant evidence.
\item 4 = mostly relevant evidence, with minor irrelevant details.
\item 5 = highly relevant evidence that directly supports the clinical prediction.
\end{itemize}

\textbf{3. temporal\_reasoning}

Does the explanation correctly reason over the event timeline?

\begin{itemize}
\item 1 = ignores or misuses temporal order.
\item 2 = mentions time but does not use it meaningfully.
\item 3 = partially uses temporal sequence.
\item 4 = mostly uses temporal progression correctly.
\item 5 = clearly reasons over progression, escalation, de-escalation, or stability over time.
\end{itemize}

\textbf{4. clinical\_coherence}

Is the reasoning medically plausible and internally consistent?

\begin{itemize}
\item 1 = clinically incoherent or contradictory.
\item 2 = weak clinical logic with major gaps.
\item 3 = partially coherent but unclear or incomplete.
\item 4 = mostly coherent and medically plausible.
\item 5 = clinically coherent, plausible, and internally consistent.
\end{itemize}

\textbf{5. completeness}

Does the explanation provide enough evidence to justify the conclusion?

\begin{itemize}
\item 1 = insufficient explanation.
\item 2 = missing several important pieces of evidence.
\item 3 = partially sufficient but incomplete.
\item 4 = mostly complete with minor omissions.
\item 5 = complete and well-supported.
\end{itemize}

\textbf{6. safety\_overclaiming}

Does the explanation avoid unsupported severity claims, diagnoses, causal claims, risk claims, or misleading wording?

\begin{itemize}
\item 1 = frequent overclaiming or potentially unsafe claims.
\item 2 = several unsupported severity or causal claims.
\item 3 = some overclaiming or unsupported wording.
\item 4 = mostly cautious, with minor wording issues.
\item 5 = cautious, appropriately qualified, and not misleading.
\end{itemize}

\textbf{7. outcome\_alignment}

Does the final prediction and explanation align with the ground-truth label?

\begin{itemize}
\item 1 = prediction/conclusion is wrong or explanation supports the wrong outcome.
\item 2 = mostly misaligned with the correct outcome.
\item 3 = partially aligned but ambiguous or weak.
\item 4 = mostly aligned with the correct outcome.
\item 5 = clearly aligned with the correct outcome.
\end{itemize}

\textbf{8. clinical\_usefulness}

Would this explanation help a clinician understand the prediction?

\begin{itemize}
\item 1 = misleading, unsafe, or clinically unhelpful.
\item 2 = weak usefulness; may confuse the reader.
\item 3 = somewhat useful but incomplete or partly misleading.
\item 4 = clinically useful with minor limitations.
\item 5 = highly useful, well grounded, and decision-relevant.
\end{itemize}

\textbf{Task-specific rules}
\begin{itemize}
\item Use the task instruction to judge whether evidence is relevant.
\item For length-of-stay prediction, relevant evidence should reflect care intensity, clinical complexity, escalation or de-escalation, procedures, monitoring burden, medication burden, complications, or stability.
\item Do not over-reward routine vitals, routine labs, isolated medications, or copied event lists unless the explanation clearly connects them to the task.
\item Penalize unsupported claims such as ``severe'', ``high-acuity'', ``routine'', ``low-risk'', ``major complication'', ``no complication'', ``prolonged QT'', or ``organ failure'' unless the provided events support that interpretation.
\item Absence-based claims such as ``no escalation'', ``no organ failure'', or ``no major complication'' are allowed only when stated cautiously and based on the observed event timeline.
\item Patient age, prediction time, or demographic information may be used only if explicitly provided in the input.
\item Penalize reasoning steps that cite Evidence numbers but draw conclusions not supported by those evidence items.
\item Penalize the explanation if the Conclusion line is missing, inconsistent with the model prediction, or inconsistent with the preceding reasoning.
\end{itemize}

\textbf{Important rules}
\begin{itemize}
\item Use only the provided clinical events and patient context as evidence.
\item Do not reward fluent writing if it is not grounded.
\item Do not reward a correct prediction if the explanation is unsupported.
\item If the model prediction is wrong, outcome\_alignment should be low.
\item If the explanation supports the wrong clinical conclusion, clinical\_usefulness should be low.
\item Return only valid JSON.
\item Do not wrap JSON in markdown.
\end{itemize}

\end{promptbox}

\subsection{Artifact and Data Usage}

EHRSHOT is used under its corresponding data use agreement and access restrictions for research purposes only. ChatHealthAI is intended solely for research and experimental evaluation, and this work does not publicly release processed patient data or unrestricted clinical artifacts.

\subsection{Computational Resources}

ChatHealthAI uses CLMBR-T-Base together with DeepSeek-R1-Distill-Qwen-14B as the frozen LLM backbone. Clinical event refinement and teacher-generated reasoning supervision are produced using GPT-oss-120B. Experiments were conducted on NVIDIA RTX 6000 Pro Blackwell 96GB GPUs.

\subsection{Implementation Details}

Our implementation is based on PyTorch and HuggingFace Transformers. CLMBR-T-Base is loaded through the FEMR framework. Evaluation metrics are computed using scikit-learn. Training uses the AdamW optimizer with learning rate $1\times10^{-5}$ and weight decay $0.01$. Mixed-precision bfloat16 training is used.

\end{document}